\documentclass[journal]{IEEEtran}
\usepackage[utf8]{inputenc}
\usepackage{cite}
\usepackage{graphicx}
\usepackage{subfigure}
\usepackage{amsmath,amssymb,amsfonts}
\usepackage{color}
\usepackage{bm}
\usepackage{algorithm}
\usepackage{algorithmic}
\usepackage{etoolbox}
\usepackage{enumerate}
\usepackage{amsthm}
\usepackage{multirow}
\usepackage{mathrsfs}
\usepackage{url}
\usepackage{array}

\allowdisplaybreaks[4]

\makeatletter
\patchcmd{\@makecaption}
  {\scshape}
  {}
  {}
  {}
\makeatletter
\patchcmd{\@makecaption}
  {\\}
  {.\ }
  {}
  {}
\makeatother

\makeatletter
\long\def\@makecaption#1#2{%
  \vskip\abovecaptionskip
  \sbox\@tempboxa{\normalfont\footnotesize #1: #2}%
  \ifdim \wd\@tempboxa >\hsize
    {\normalfont\footnotesize #1: #2\par}
  \else
    \global \@minipagefalse
    \hb@xt@\hsize{\box\@tempboxa\hfil}%
  \fi
  \vskip\belowcaptionskip}
\makeatother

\newcommand{\rtwo}[1]{\textcolor{blue}{#1}}

\renewcommand{\figurename}{Figure}
\def\tablename{Table}

\begin{document}

\title{MSS-DepthNet: Depth Prediction with Multi-Step Spiking Neural Network}

\author{Xiaoshan Wu, Weihua He, Man Yao, Ziyang Zhang, Yaoyuan Wang, Guoqi Li\dag,~\IEEEmembership{Member,~IEEE}
\thanks{X. Wu is with ZJU-UIUC Institute, Zhejiang University, Haining, Zhejiang, China.}
\thanks{W. He is with Department of Precision Instrument, Tsinghua University, Beijing, China.}
\thanks{M. Yao is with School of Automation Science and Engineering, Xi'an Jiaotong University, Xi'an, Shaanxi, China.}
\thanks{Z. Zhang and Y. Wang are with Advanced Computing and Storage Lab, Huawei Technologies Co. Ltd, Beijing, China.}
\thanks{G. Li is with Institute of Automation, Chinese Academy of Sciences, Beijing, China.}
\thanks{\dag: Corresponding author,  \protect\url{guoqi.li@ia.ac.cn}.}}
\maketitle

\begin{abstract}
Event cameras are considered to have great potential for computer vision and robotics applications because of their high temporal resolution and low power consumption characteristics. However, the event streams output from event cameras have asynchronous, sparse characteristics that cannot be handled by existing computer vision algorithms. Spiking neural network is a novel event-based computational paradigm that is considered to be well suited for processing event camera tasks. However, direct training of deep SNNs suffers from degradation problems. This work addresses these problems by proposing a spiking nerual network architecture with novel residual block designed and multi-dimension attention modules combined, focusing on the problem of depth prediction. In addition, a novel event stream representation method is proposed specifically for SNNs. This model outperforms previous ANN networks of the same size on the MVSEC dataset and shows great computational efficiency.
\end{abstract}


\section{Introduction}\label{sec:Intro}
The human brain can efficiently perform highly complex tasks with only about 20 Watts of power consumption by transmitting information in the form of spikes between neurons. Traditional artificial intelligence neural networks (ANNs) simulate the human brain at the cost of intensive computation, heavily relying on hardware computing power. However, in the post-Moore era, the problem that the increase in computing power cannot keep up with the demand for algorithm development has become the major bottlenecks limiting the development of traditional Artificial Intelligence (AI). Brain-like computing is a new computing paradigm inspired by the way the biological nervous system processes information, and is considered to be one of the most important breakthroughs to disrupt the existing intelligent computing technology. In recent years, many researches on chip architecture, computational theory, and model algorithms around brain-like computing have emerged. Among them, the spiking neural network (SNN) is considered the core of brain-like computing research, and has attracted the attention of many scholars with its rich neurodynamic properties in the spatio-temporal domain and event-driven feature. The utilization of SNN was largely limited until the direct training method of SNNs was proposed. Wu et al. proposed a spatio-temporal backpropagation algorithm for SNN\cite{wu2018spatio}.Wu et al. further proposed a direct training method on the basis of the back propagation algorithm \cite{wu2019direct}. On the basis of this, algorithms on direct training spiking neural network have flourished. For example, Deng et al. further proposed a gradient re-weighting method for SNN algorithm \cite{deng2022temporal}.This lays a foundation for expanding the application of SNN.

Originated from brain science, SNN uses spikes with precise discharge time as the basic carrier of operations. Due to this feature, SNNs needs large-scale datasets with rich real event-based spatio-temporal information. Inspired by bio-vision processing mechanism, event-based camera is a novel neuromorphic visual sensor that capture the brightness change information in each pixel on the visual field and transfer this temporal information into an asynchronous stream of event. The low latency, high dynamic range, and low energy consumption characteristics of event-based cameras make them very promising in many fields, such as optical flow estimation, motion recognition, and depth prediction. It is believed that when processing event camera datasets, SNNs, which are also event-based, are inherently more suitable than traditional ANNs. Since SNNs can exploit the sparse nature of event-based cameras to reduce energy consumption, while traditional ANNs will perform unnecessarily intensive operations ignoring the sparse nature of event camera datasets. Meanwhile SNN can process events in a timely manner to further reduce latency. The direct processing of event streams using SNNs is becoming a research hotspot and has yielded good results in some low-level visual tasks such as trajectory prediction \cite{guiji}, optical flow estimation \cite{guangliu}, and angular velocity estimation \cite{jiaosudu}. 

In terms of applying SNNs to event cameras, the vision tasks handled by SNNs gradually move from simple, low-level vision tasks \cite{GarrickOrchard2015HFirstAT} with unsupervised learning, to more advanced task by indirect supervised learning, i.e., converting ANN models into SNNs \cite{Connor2013}, to complex tasks with SNN training models directly \cite{StereoSpike}. However, existing models trained directly with SNNs either fail to utilize stateful SNNs structure \cite{StereoSpike} or compensate for the network degradation in SNNs through a hybrid ANN-SNN co-training network \cite{SpikeFlowNet}. The application of direct training algorithms to train multi-step SNNs to process event camera datasets remains to be explored. Multi-step SNNs can handle pulse sequences with multiple consecutive moments of input, and they have more information in time dimension than single-step SNNs. In the biological vision system, a series of images with parallax generated by the motion of an organism at a certain time is an important basis for biological processing of visual information. We believe that it is more biologically reasonable to upgrade the stateless single-step SNN to the stateful multi-step SNN.

\begin{figure}
\centering
\includegraphics[width=0.5\textwidth]{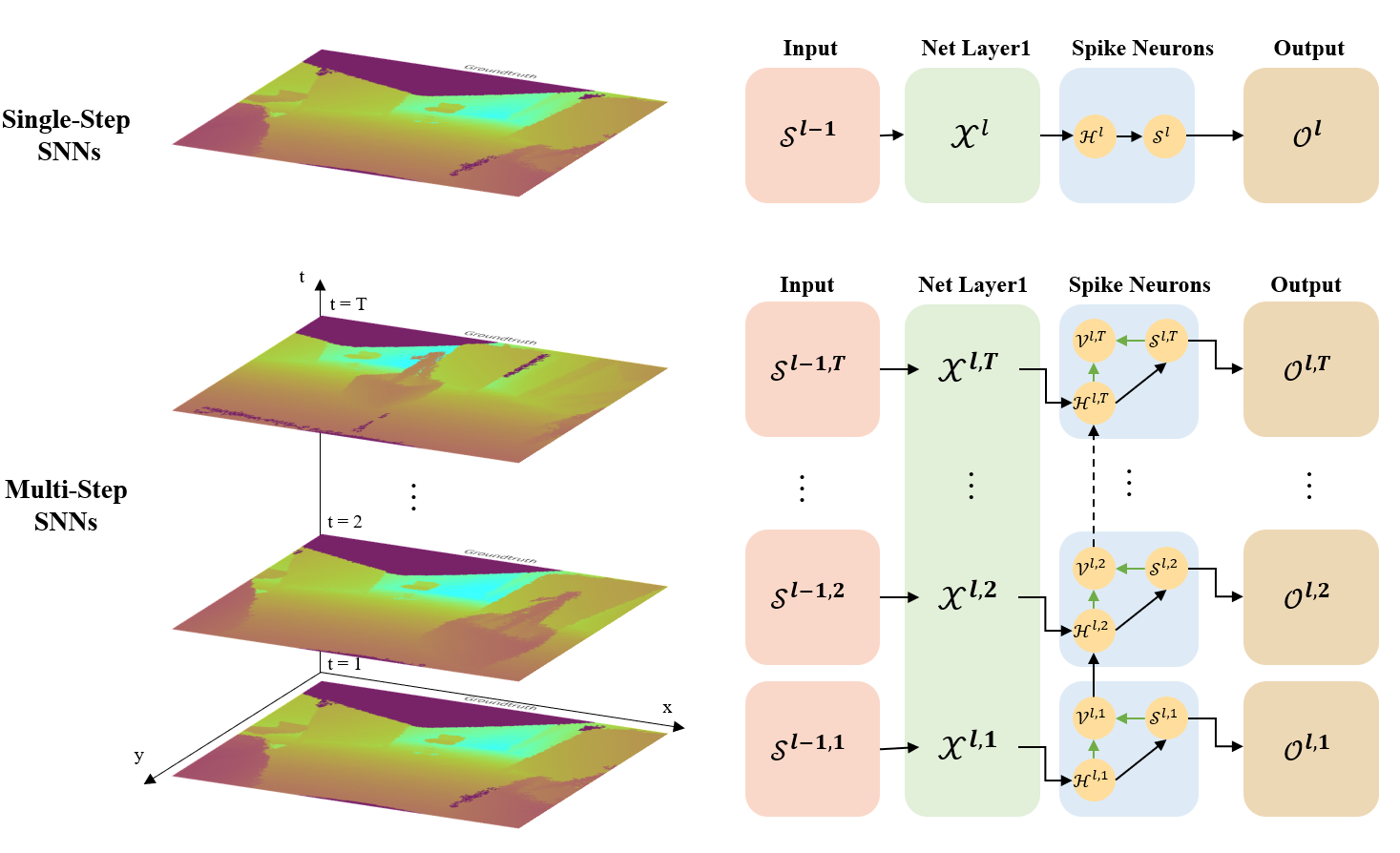}
\caption{Single Step SNNs structure Vs Multi-Step SNN structure in network layer $l$. In single-step SNNs, the input is from certain time stamp. The spike neuron is used as filter for only input charge from last layer. In multi-step SNNs, the input contains a series of data from a continuous time period $[1,2,...,T]$, which are fed into the net at the same time. The spike neuron not only receive information from last layer, but also affected by the membrane potential from previous neuron. In this way, additional temporal information is included.}
\label{multi-step}
\end{figure}

To meet this goal, this work applies the Attention Residual learning method, which optimizes the direct training SNN algorithm, to the event camera task, and applies it to the specific problem of depth estimation. To summarize, the main contributions of this work are as follows: 

\begin{itemize}
    \item \rtwo{The research in this paper addresses the bottleneck that constrains the application of multi-step deep SNNs to stereo scenes and provides a new technical means for SNNs to handle event camera tasks. }
    \item \rtwo{We put up with Cumulative Stacking method for event data representation.}
    \item \rtwo{This work also validates the rationality of Attention Residual Block in SNNs by its application on depth estimation.}
\end{itemize}


\section{Related Works}\label{sec:Relate}

\rtwo{The advent of event cameras provides new solutions for tasks applied to visual scenarios but also presents the challenge of designing novel algorithms applicable to asynchronous and sparse event streams. SNNs are applicable to event camera datasets but suffer from the inability to outperform ANNs. This subsection first reviews and discusses deep learning algorithms including SNNs applied to event camera datasets, and then synthesizes and evaluates two of the more popular approaches to optimize SNNs at the moment.}

\subsection{SNNs in Event Camera Task}

SNN was first applied to target recognition, action classification tasks using unsupervised learning for simple scenes in embedded devices \cite{EventSurvey}. Orchard et al. \cite{GarrickOrchard2015HFirstAT} processed Address Event Representation (AER) vision sensor output for target recognition using a spike hierarchy model incorporating a Gabor filter for extracting features and a classifier for recognizing spike signal patterns. This represents a breakthrough in computational vision models from synchronous to asynchronous, and SNN is more suitable for processing this new model. With the rise of deep learning in computer vision, some scholars have tried to convert trained ANNs into SNNs that can operate directly on event data as a way of indirect supervised learning for more complex tasks \cite{EventSurvey}. O'Connor et al. converted the model to an event-driven SNN network for MNIST handwritten digit recognition after training Deep Belief Networks (DBNs) offline \cite{Connor2013}. However, the limitation of these models transformed from ANN to SNN is that they can only be applied to shallow networks, which does not work well when dealing with complex tasks that apply large data such as Multi Vehicle Stereo Event Camera (MVSEC), so scholars have gradually turned their attention to training SNN directly \cite{EventSurvey}\cite{SpikeFlowNet}. Spike-FlowNet [10] uses a hybrid ANN-SNN neural network structure to estimate the optical flow at the output of sparse event cameras, outperforming ANN-based methods while offering significant computational efficiency. This again demonstrates the effectiveness of SNN in handling sparse and asynchronous event camera datasets. StereoSpike \cite{StereoSpike} completely uses the SNN network for depth estimation, but its performance cannot completely exceed that of ANN, while the time step in SNN is set to 1, which fails to fully exploit the characteristics of SNN time dimension. There is still a gap in the field of processing event camera datasets entirely using multi-step SNN networks.

\subsection{Attention Residual Learning of SNNs}
SNNs are an ideal model for processing event camera output, but the lack of SNN direct training algorithms makes SNNs often not as effective as ANNs. Direct training of deep SNNs has been the hard part of the SNN algorithm field, and is considered to be the door knocker that can really open the door of the deep learning community. One of the reasons for this difficulty is the problem of disappearing spikes in the deeper layers of the SNNs. And considering the special spike information form of SNN, it is not feasible to directly transplant the proven techniques of coping with deep neural networks in ANN networks to SNNs. To overcome these problems, a residual block structure suitable for SNNs was proposed by Hu et al \cite{MS-ResNet}. Yao et al. \cite{ManYao2022AttentionSN} further proposed a plug-and-play multi-dimension attention block, including temporal, spatial and channel blocks specialized for SNNs. This attention block can help with reducing unnecessary spikes while improving performance. However, it remains to be seen how such plug-and-play attention modules and residual networks will work together in specific SNN tasks.


\section{Method}\label{sec:Method}

\subsection{Cumulative Stacking Event Representation Protocol}\label{subsec:Cumulative Stacking}
The Multi Vehicle Stereo Event Camera (MVSEC) dataset is used in this work, which is collected from car, motorcycle, hexacopter and handheld data and fused with LiDAR, IMU, motion capture and GPS to provide ground truth attitude and depth imagery \cite{AlexZihaoZhu2018TheMV}. Specifically, indoor\_flying sequence recorded by a drone in the lab environment is used in this work. The event camera one the drone can track the light intensity of the image. When the logarithmic intensity on a pixel changed above the set threshold, the camera will return the pixel coordinates, timestamp and the brightness changed direction, i.e., a tuple of (x, y, t, p). Furthermore, there are two Dynamic and Active Pixel Vision Sensor (DAVIS) on the left and right sides of the drone separately, providing binocular vision. 

\begin{figure*}
\centering
\includegraphics[width=1\textwidth]{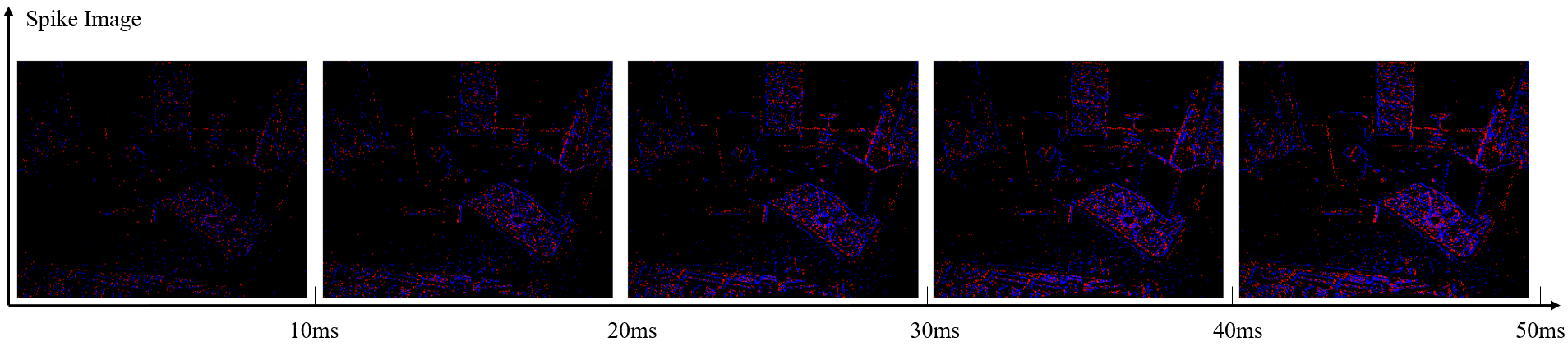}
\caption{This image shows the cumulative stacking process of T = 5 in 50 ms. The 5 frames used for the input spike accumulation is collected from 0~10 ms, 0~20 ms, …, 0~50 ms separately.}
\label{input}
\end{figure*}

Due to the asynchronous and sparse characteristics of event-based camera dataset, special event representation protocol needs to be utilized. Wang et al. \cite{LinWang2018EventbasedHD} proposed two methods, Stacking Based on Time (SBT) based on spikes accumulated in certain time window, and Stacking Based on the number of Events (SBE) based on spikes accumulated when event number exceeds a certain threshold. The second method will lose the time information during the accumulation process, so we utilize the idea of the first method and make improvement based on it, accumulating the spikes cumulatively in the time window of 50 ms at each pixel. To avoid the cancellation between increment polarity and decrement polarity, this work uses two channels for different polarities. The length of the time window needs to be chosen carefully, because if the time window is too small, the events will be too sparse and the model will not be able to extract effective information, and if the time window is too large, the events will be piled up and useful information will also be lost. Experiments showed that using a time window of 50 ms to accumulate the pulses could maximize information utilization. Note that the ground truth returned from LIDAR is also provided at the frequency of 20 Hz. To construct the input for multistep SNNs, we use the cumulative stacking method. That is, if time step in SNN is chosen as T, then the 50 ms would be divided into T frames. For each 50 ms, the cumulation for each frame is on the basis of previous frame. Figure \ref{input} shows an example of T=5. In this way, the (x, y, t, p) original data is transformed into the form of [T, C, H, W], where T is SNN time step, C is channel number, 2 for monocular input and 4 for binocular input, H for height and W for width.

\subsection{SNN models}\label{subsec:IF}

\begin{figure}
\centering
\includegraphics[width=0.5\textwidth]{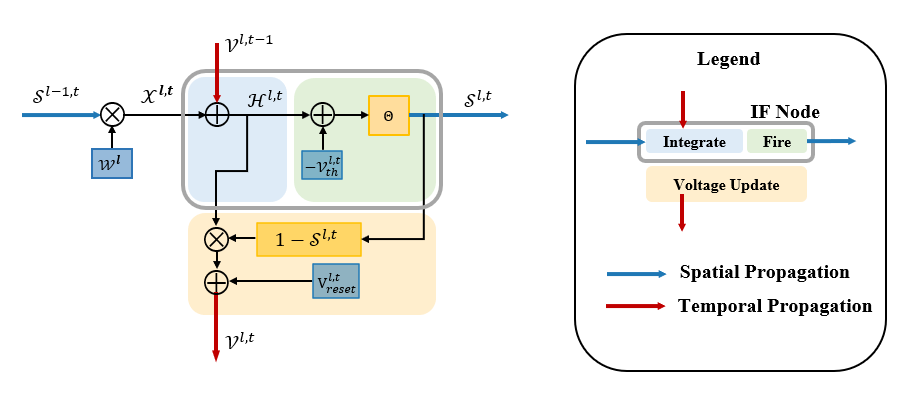}
\caption{Structure of IF node. The left side is the vector diagram showing the internal neural dynamics of the IF node, which is in corresponding with Eq \ref{IF_Eq}. The right side is the conceptional diagram showing how temporal and spatial information are forwarded in IF node. }  
\label{IF}
\end{figure}

The fundamental difference between a spiking neuron network and a normal artificial neural network lies in the construction of neurons. Unlike traditional ANNs neuron, which is entirely dependent on spatial domain input, spiking neuron functions on the spatio-temporal domain. To utilize the neural dynamics characteristic of neuromorphic computing, the Integrate-and-Fire model is used to implement spiking neurons.

\begin{equation}\label{IF_Eq}
\begin{aligned}
&\bm{\mathcal{X}}^{l,t} = \sum_{i=1}^{T}{\bm{{w}}^{l,i} \bm{\mathcal{S}}^{l-1,i}} \\
&\bm{\mathcal{H}}^{l,t} = \bm{\mathcal{V}}^{l,t-1} + \bm{\mathcal{X}}^{l,t-1} \\
&\bm{\mathcal{S}}^{l,t} = {\bm{\Theta}(\bm{\mathcal{H}}^{l,t} - \bm{\mathcal{V}}_{th}^{l,t})} \\
&\bm{\mathcal{V}}^{l,t} = \bm{\mathcal{H}}^{l,t} \cdot (1-\bm{\mathcal{S}}^{l,t})+\bm{\mathcal{V}}_{th}^{l,t} \cdot \bm{\mathcal{S}}^{l,t}  \\
\end{aligned}
\end{equation}

where ${\bm{\mathcal{X}}^{l,t}}$ is the weighed spike accumulation in layer $l$ at the $t^{th}$ IF node. The membrane potential before the spike, ${\bm{\mathcal{H}}^{l,t}}$, is the summation of the current membrane potential ${\bm{\mathcal{V}}^{l,t-1}}$ and external input ${\bm{\mathcal{X}}^{l,t}}$. ${\Theta}$ is the step function. If the membrane potential ${\bm{\mathcal{H}}^{l,t}}$ surpasses the neuron voltage threshold, the neuron will fire the spike and the membrane potential will be set to ${\bm{\mathcal{V}}^{l,t}_{reset}}$. Otherwise, the neuron won’t spike and keep the current membrane potential. This structure of the model is illustrated in Figure \ref{IF}. 

\subsection{Net Architecture}
As a commonly used architecture in semantic segmentation, which shares similar contextual information with depth prediction, a four-layer U-Net \cite{OlafRonneberger2015UNetCN} is chosen as a backbone for depth prediction in this work, which is composed of encoder, residual block and decoder. In encoder, depth feature is captured by contraction path layer-by-layer. The extracted depth features then go through an upsampling path in decoder layer, in which the depth features would be combined with high resolution features from symmetric path propagation to reduce the loss of detailed information. The output from each decoder block has two uses, one for depth prediction at each layer, and another for the input to the attention block at the next layer. Additionally, previous work \cite{StereoSpike} shows that nearest neighbor (NN) is preferable to linear interpolation for SNN since the latter would produce non-integer values, which is against the philosophy of SNN. Therefore, the nearest neighbor is used for upsampling here. Two residual blocks are added to better retain feature mappings in deep neural networks. On the basis of traditional U-Net structure, the temporal-channel-spatial attention block is added as a general component. 

\begin{figure*}
\includegraphics[width=1.0\textwidth]{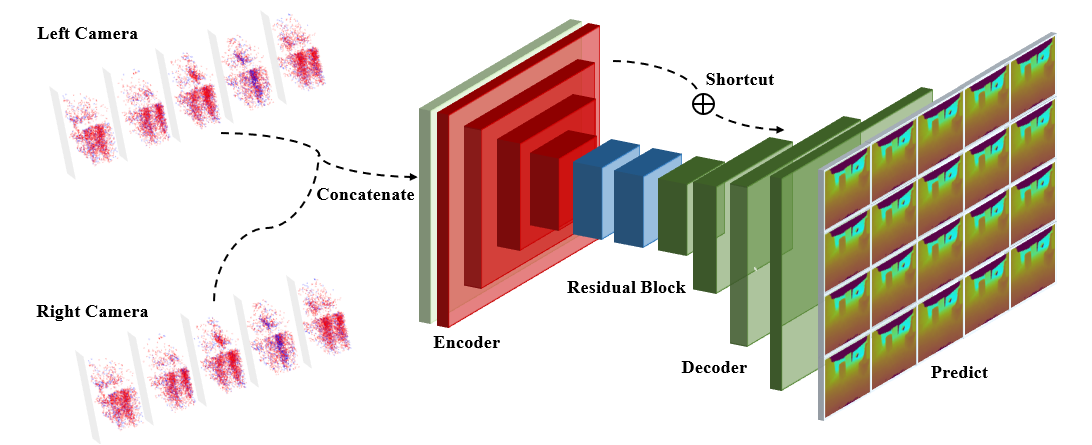}
\caption{Overall architecture of the model. The input of the model is concatenated by spike cumulative staking from left and right cameras. The input spikes the go through a four-layer encoder, two residual blocks and a four-layer decoder in order. Note that in decoder, the output of each layer will be summed together with its symmetry part from encoder as an input fed to the next layer in decoder. Each neuron in each layer in decoder would produce a prediction. In this four-layer model with a time step of five, 20 different prediction would be produced. Normally we choose the prediction from the last neuron in the last layer to compare with the ground truth. The specific structure of encoder, residual and decoder blocks are illustrated in Figure. \ref{blocks}. } 
\label{3D_Net}
\end{figure*}

\subsubsection{Attention block}
A multi-dimension attention block specialized for SNNs is used in this work. Woo et al.\cite{SanghyunWoo2018CBAMCB} proposed the channel and spatial attention blocks. Yao et al. \cite{ManYao2022AttentionSN} further generalized the attention module of the channel dimension to the time dimension. In this work, the attention block is composed of three attention module in a roll, i.e temporal attention module, channel attention module and spatial attention module. Note that each module is a plug-and-play block, meaning that we can selectively use the attention module we want to use. For example, in Figure \ref{blocks}, we only use the channel attention module and the spatial attention module, i.e the CSA block. In each layer in the MSS model, the attention block is placed before contraction in encoder block, and after upsampling at decoder block, aiming at extracting more detailed information. In this section, we illustrated the most generally attention structure TCSA. 

\begin{figure}
\centering
\includegraphics[width=0.5\textwidth]{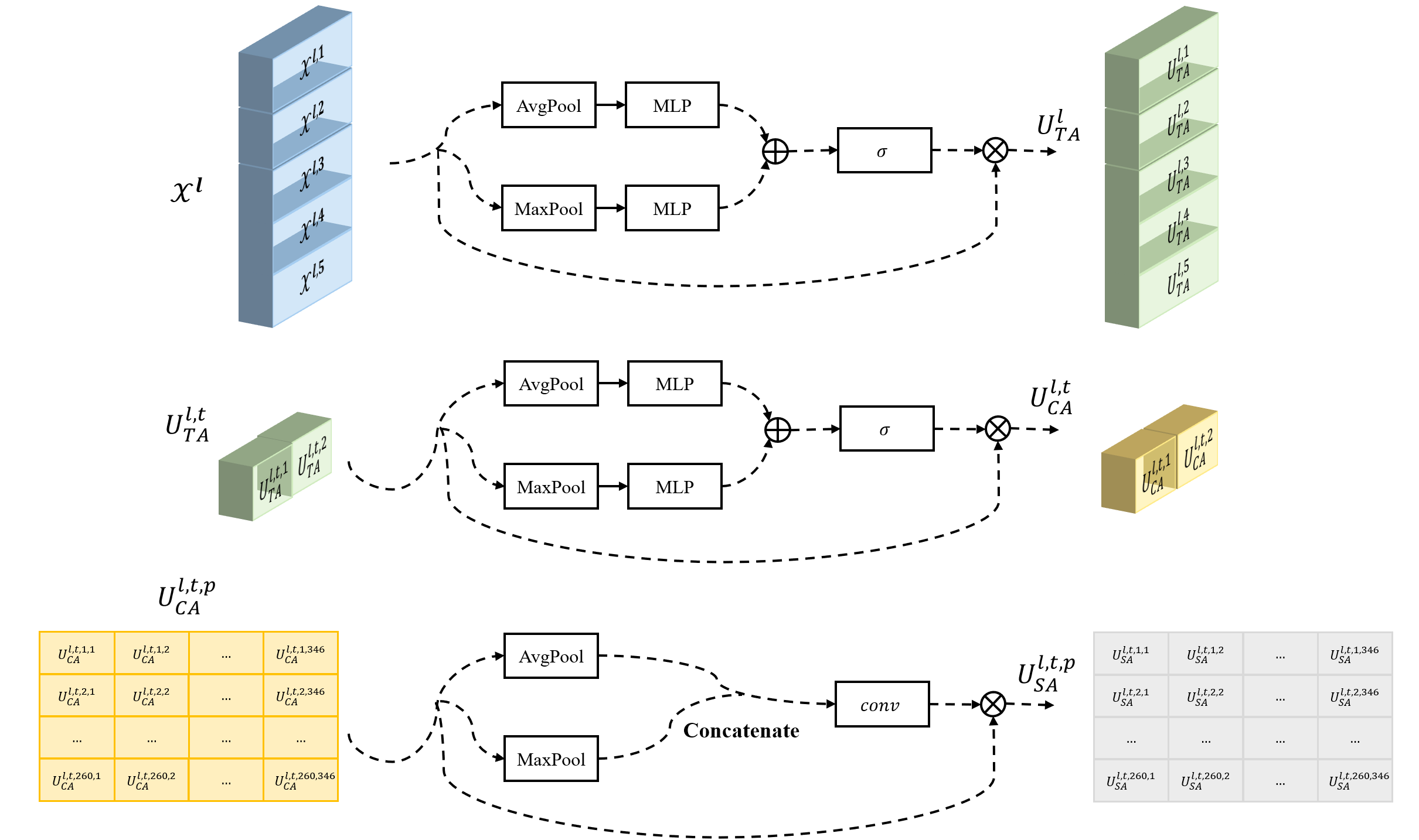}
\caption{Data propagation in a general Temporal-Channel-Spatial Attention (TCSA) block. In this example. Input $\bm{X}^{l}$ is in shape of $[T=5, C=2, H=260, W=346]$. Each layer is expanded from previous layer.}  
\label{Att}
\end{figure}
For temporal attention module, the input in layer $l$ is $\bm{X}^{l}=[\bm{\mathcal{X}}^{l,1},\bm{\mathcal{X}}^{l,2},...,\bm{\mathcal{X}}^{l,T}]$. The shape of $\bm{X}^{l}$ is $[T, C, H, W]$ and $\bm{X}^{l}$ is used in two branches. Each branch is composed of a 1-D pooling layer for aggregating channel-spatial information and a $MLP$ layer to obtain a ${1\times1\times T}$ shape attention map. Pooling method used here are max pooling and average pooling. In $MLP$ layer, parameter number  can be lessen from ${T}$ to ${T/r}$, where $r$ is parameter reduction factor in hidden layer. The final temporal attention map ${U_{TA}^{l}}$ is obtained by activating the sum of corresponding outputs from $MLP$s 

\begin{equation}\label{TA}
\begin{aligned}
\bm{g}_{TA}(\bm{X}^{l}) =& \bm{\sigma}({MLP}({AvgPool}(\bm{X}^{l})) \\
& + {MLP}({MaxPool}(\bm{X}^{l}))) \\
\bm{U}_{TA}^{l}(\bm{X}^{l}) =& \bm{g}_{TA}(\bm{X}^{l}) \otimes \bm{X}^{l} \\
\end{aligned}
\end{equation}

Channel-attention module is just like temporal-attention, except that the granularity of attention operations is smaller, shifting from a temporal dimension to a channel dimension at each time step $t$. It can be expressed as followed.

\begin{equation}\label{CA}
\begin{aligned}
&\bm{U}_{CA}^{l,t}(\bm{U}_{TA}^{l,t}) = \bm{g}_{CA}(\bm{U}_{TA}^{l,t}) \otimes \bm{U}_{TA}^{l,t} \\
\end{aligned}
\end{equation}

For spatial-attention module, the granularity further shrink from channel dimension to spatial dimension at each channel $p$. There are still two branches, each branch is composed of a 2-D pooling layer, either max pooling or average pooling. The results are concatenated and then fed into a convolution layer with a filter of size $3\times3$.

\begin{equation}\label{SA}
\begin{aligned}
&\bm{g}_{SA}(\bm{U}_{CA}^{l,t,p}) = \bm{conv}([{AvgPool}(\bm{U}_{CA}^{l,t,p}) ,{MaxPool}(\bm{U}_{CA}^{l,t,p})]) \\
&\bm{U}_{SA}^{l,t,p}(\bm{U}_{CA}^{l,t,p}) = \bm{g}_{SA}(\bm{U}_{CA}^{l,t,p}) \otimes \bm{U}_{CA}^{l,t,p} \\
\end{aligned}
\end{equation}
The data propogation process is illustracted in Figure.\ref{Att}.

\subsubsection{Residual block}
\begin{figure}
\includegraphics[width=0.5\textwidth]{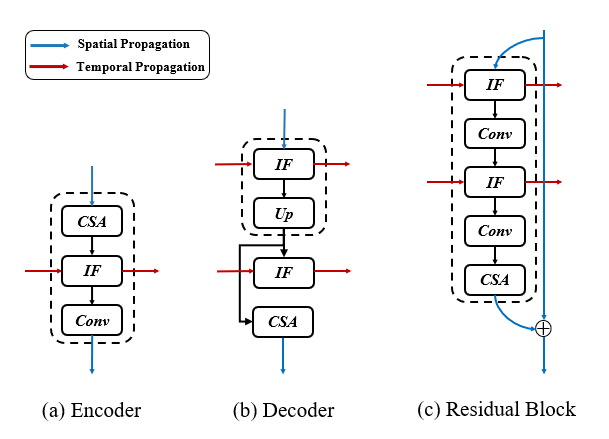}
\caption{Spatio-temporal propagation in model. (a) Structure of each layer of encoder. (b) Structure of each layer of decoder. The output of upsampling is used for both depth prediction in IF node and channel-spatial dimension attention. (c) Residual block structure. Note that temporal attention module is not shown in this graph since it is operated on a full temporal scale.} 
\label{blocks}
\end{figure}

Hu et al. showed that applying classic ResNet designed for the ANN version, like Vanilla ResNet, to SNN mechanically is ineffective \cite{MS-ResNet}. Therefore, this work utilized the basic structure of MS-ResNet, which is specially designed for SNN. However, Batch Norm (BN) block after convolution is removed and and attention block is added at the very end. Two such residual blocks are used in series. In Figure \ref{blocks}.c, the internal structure of residual block is shown to explicit demonstrate the internal neural dynamics of the structure.
\subsubsection{loss}
There are previous works \cite{SpikeFlowNet} and \cite{StereoSpike} utilized the scale-invariant and shift-invariant loss function proposed by Ranftl et al. \cite{ReneRanftl2020TowardsRM} works well for event-camera based tasks. Here, we followed this tradition. The predicted depth for a single frame accumulated on time window $k$ is ${\bm{\mathcal{D}}_{k}}$ is the membrane potential of the last predict IF node from the last decoder block of the U-Net, since it contains the richest temporal-channel-spatial information. The difference between the ground truth ${\bm{\mathcal{\hat{D}}}_{k}}$ and predicted depth ${\bm{\mathcal{D}}_{k}}$ is $\bm{\mathcal{R}}_{k}=\bm{\mathcal{\hat{D}}}_{k}-\bm{\mathcal{D}}_{k}$.

Then the he scale-invariant and shift-invariant loss function for a frame $k$ is expressed as
\begin{equation}\label{loss_ssi}
\begin{aligned}
& \bm{\mathcal{L}}_{k,ssi} = \frac{1}{n} \times \sum_{u}(\mathcal{R}_{k}(u))^2+\frac{1}{n^2} \times (\sum_{u}(\mathcal{R}_{k}(u)))^2
\end{aligned}
\end{equation}
, where $u$ represent the valid ground truth pixels and $n$ is the total number of u. Another regulation loss is used to penalize the derivative in the spatial direction $x$ and $y$ for being too large, which is defined as
\begin{equation}\label{loss_reg}
\begin{aligned}
& \bm{\mathcal{L}}_{k,reg} = \frac{1}{n} \times \sum_{u}\left|\nabla_{x}\mathcal{R}_{k}(u)\right|+\left|\nabla_{y}\mathcal{R}_{k}(u)\right|
\end{aligned}
\end{equation}
The total loss function is expressed as the weighted summation of the scale-shift-invariant loss and the regulation loss. The total loss is the summation of loss on all input frame.
\begin{equation}\label{loss}
\begin{aligned}
& \bm{\mathcal{L}} = \sum_{k}(\bm{\mathcal{L}}_{k,ssi}+\lambda\bm{\mathcal{L}}_{k,reg})
\end{aligned}
\end{equation}
\subsubsection{Training Details}
With the introduction of the surrogate gradient method, it is possible to directly use the gradient descent method for SNN training. SpikingJelly \cite{SpikingJelly} is an open-source framework that is compatible with machine learning framework and spike network simulation framework. We use it to train our spiking neural network. In this work, we used multistep IF nodes for SNNs, utilized the regular backpropagation and reset the membrane potential for each neuron to zero for each batch train cycle. According to \cite{StereoSpike}, we used the Adam optimizer with default beta coefficients and choose a learning rate of 0.002 without weight decay. We will halve the learning rate for certain milestones.


\section{Experiments}\label{sec:Exp}
The experiments in this work is implemented using PyTorch and SpikingJelly \cite{SpikingJelly}
\subsection{Performance}
We used the indoor\_flying dataset from MVSEC datasets \cite{AlexZihaoZhu2018TheMV}, and followed the same data split method as \cite{tulyakov2019learning} to divide the data into three sequences, each has separate training, validation, and test sets. We also removed the take-off and landing frames for a fair comparison with previous work. We used \itshape Mean Depth Error (MDE) \upshape as evaluation criterion.

\begin{table}
\caption{MDE values calculated by different models using indoor\_flying as the dataset. For a fair comparison, all models are using the same input preprocessing protocol proposed by \cite{tulyakov2019learning}, utilizing the same split methods and removing take-off and landing frames. All results were trained using binocular data, i.e., receive event stream from both left and right cameras.}
\label{Table_all_compare}
\centering
\setlength{\tabcolsep}{10pt}
\begin{tabular}{l l l l l}
\hline
\multicolumn{2}{c}{\multirow{1.2}*{Method}} &
\multicolumn{3}{c}{\multirow{1.2}*{MDE (cm)}}\\
\hline
\multirow{1.2}*{Model} & \multirow{1.2}*{Net} & \multirow{1.2}*{split1} & \multirow{1.2}*{split2} & \multirow{1.2}*{split3}  \\
\hline
Ours & SNN & \textbf{16.5} & \textbf{25.4} & \textbf{18.7} \\
StereoSpike  \cite{StereoSpike} & SNN & 18.7 & 29.4 & 25.4 \\
DDES \cite{DDES} & ANN & 16.7 & 29.4 & 27.8 \\
TTES \cite{TESE} & ANN & 36 & 44 & 36 \\
CopNet \cite{CopNet} & ANN & 61 & 100 & 64 \\
\hline
\end{tabular}
\end{table}

\begin{figure}
\includegraphics[width=0.5\textwidth]{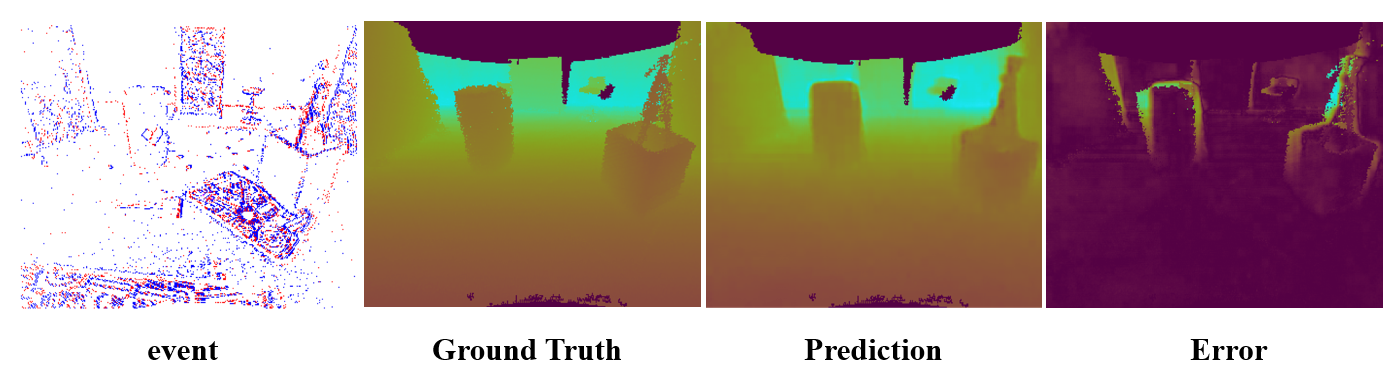}
\caption{Visualization of input spikes, ground truth from LIDAR, prediction from our model and the error between ground truth and prediction. The frame used is indoor\_flying1 frame 125.} 
\label{result}
\end{figure}

In the test, multi-step IF neurons with time step 5 were used for training and depth estimation of the event streams from the left and right binocular cameras for the MVSEC dataset indoor flight 1, 2, and 3 sequences. Table 1 gives a comparison of the MDE evaluation results with previous work on depth estimation using the same dataset. Our results show that the model performance of MSS-Depth on each set of indoor flight sequences is better to the performance of previous networks, including ANN and SNN. Note that the models in the first three line of Table \ref{Table_all_compare} are results from dense stereo prediction, meaning that ever pixel on the image is predicted no matter there is an event on the pixel or not. The models in the last two line, on the other hand, are results from sparse stereo prediction. Our work is entirely based on spiking neural network architecture, yet our result outperforms the current state-of-art ANN model DDES \cite{DDES} dealing with depth prediction to our knowledge, which is an ANN-based model. There is previous work \cite{StereoSpike} attempts to predict dense stereo depth using spiking neural network for event camera input. However, it is the first time that the SNN model outperforms the ANN model in the task of depth prediction to our knowledge. The result is visualized in Figure \ref{result}.

\subsection{Ablation Experiment}
\begin{figure}
\includegraphics[width=0.5\textwidth]{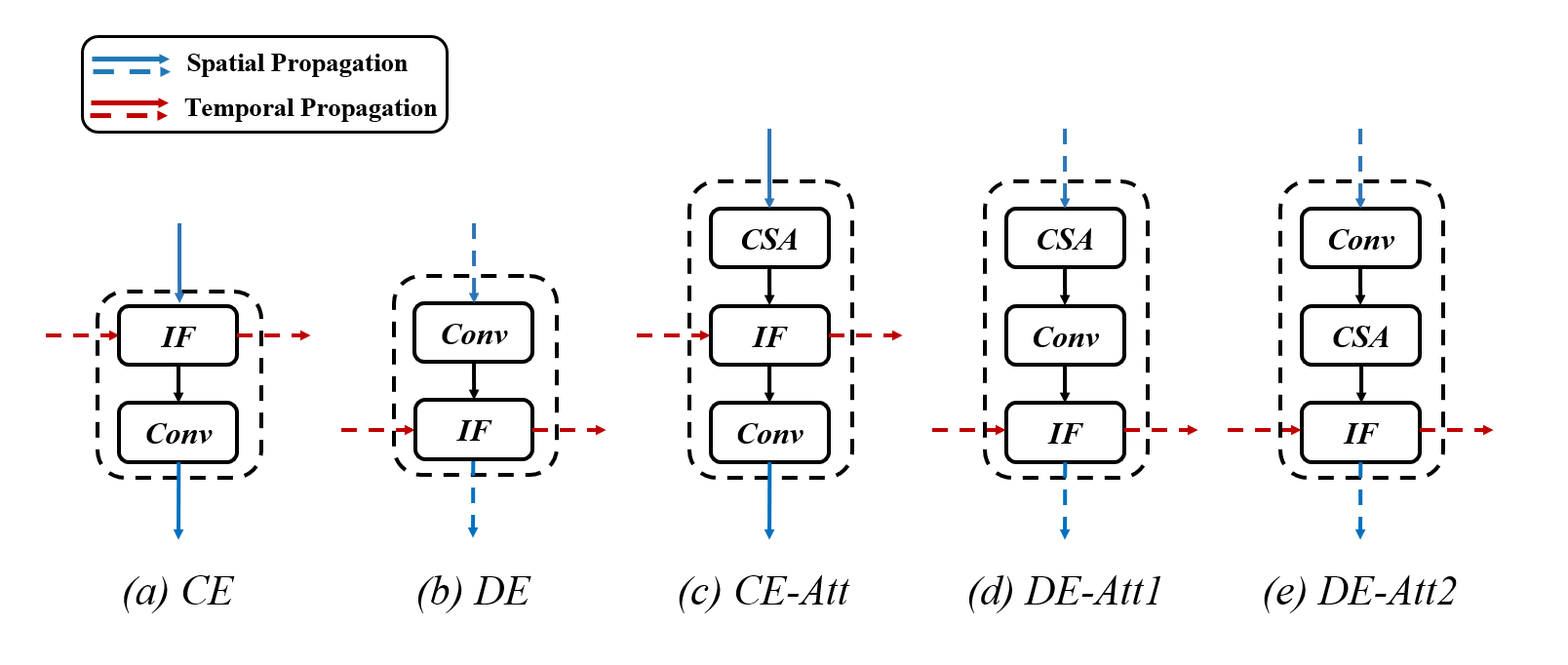}
\caption{Structure of different encoder blocks. The dashed line means that discrete values are spread in the encoder and the solid line means continuous values are used instead.} 
\label{encoder_pos}
\end{figure}

Several ablation experiments were done to determine the most appropriate position of the IF node and attention blocks. We chose the model proposed by Ulysse \cite{StereoSpike} as the baseline model. For the encoder block architecture, we tried structure illustrated in Figure \ref{encoder_pos}, and the corresponding results are shown in Table \ref{Table_encoder_compare}. CE represents continuous encoder block and DE means discrete encoder block. Between the comparison of CE and DE, and comparison between CE-Att and DE-Att, it can be found that it is better to use continuous value is propagated between encoder layer. The results difference between CE and CE-Att shows that integrated channel-spatial attention module into the net would largely improve the performance. In addition, the result of DE-Att1 shows that the the performance boost from the attention module must be combined with spiking structure, i.e. using discrete spike value at input for convolution. It is also proved from the result of DE-Att2 that adding attention module after downsampling is not as good as before downsampling, since important information that the attention module could have extracted may be lost in the process of downsampling. For the same reason, attention module is added after upsamping in decoder blocks. The encoder block used in our state-of-the-art performances model is CE-Att.

We demonstrate the effectiveness of multi-step SNNs by comparing the results of $T=1,2,5$. Note that here we tried both cumulative staking method introduced in Section \ref{subsec:Cumulative Stacking} and a refined stacking based on time method that reuse $T$ times the events accumulated in the 50ms time window as the SNN input. Since the spike would be too sparse if using the first method, we use the second method to represent input data. For this reason, temporal-wise attention module is turned off. However, we believe that the cumulative staking method can be used with higher time resolution tasks. The result from the latter method is illustrated in Table \ref{Table_encoder_compare}. The result explicitly shows improvement of results with multi-step SNNs.

\begin{table}
\caption{The performance on indoor\_flying split 1 for different encoder block architecture.}
\label{Table_encoder_compare}
\centering
\setlength{\tabcolsep}{10pt}
\begin{tabular}{l l l l l l}
\hline
\multicolumn{6}{c}{\multirow{1.2}*{Encoder Blocks Structure Comparison}}\\
\hline
\multirow{1.2}*{ } & \multirow{1.2}*{CE} & \multirow{1.2}*{DE} & \multirow{1.2}*{CE-Att} & \multirow{1.2}*{DE-Att1} & \multirow{1}*{DE-Att2} \\
\hline
\multirow{1.2}*{MDE} & \multirow{1.2}*{19.32} & \multirow{1.2}*{19.42} & \multirow{1.2}*{16.5} & \multirow{1.2}*{43.20} & \multirow{1.2}*{20.39} \\
\hline
\multicolumn{6}{c}{\multirow{1.2}*{Multi-step SNN Time Step Comparison}}\\
\hline
\multirow{1.2}*{} & \multirow{1.2}*{T=1} & \multirow{1.2}*{} &\multirow{1.2}*{T=2} &\multirow{1.2}*{} & \multirow{1.2}*{T=5}\\
\hline
\multirow{1.2}*{MDE} & \multirow{1.2}*{16.74} & \multirow{1.2}*{} &\multirow{1.2}*{16.84} &\multirow{1.2}*{} & \multirow{1.2}*{16.50}\\
\hline
\end{tabular}
\end{table}

\subsection{Efficiency}
We believe the SNN structure is very suitable for processing event camera data due to its inherent firing mechanism. The data from the event camera are usually very sparse. Lee et al. \cite{SpikeFlowNet} also argued that if ANN is used to process such data, a fixed number of dense multiply-accumulate (MAC) computations will be used regardless of the sparse input. However, for SNN, it can be seen from Table \ref{Table_fire} that only about 10\% of the SNN neurons are fired, which largely increases the efficiency of the model. Another great advantage of SNN over ANN is that SNN utilizes accumulation (AC) while ANN utilizes multiply-accumulate (MAC) computation in the computation, and AC consumes much less energy than MAC operation.
\begin{table}
\caption{Spike Activity Rate for binoculars in percentage.}
\label{Table_fire}
\centering
\setlength{\tabcolsep}{10pt}
\begin{tabular}{l l l l}
\hline
\multicolumn{4}{c}{\multirow{1.2}*{Firing Rate}}\\
\hline
\multirow{1.2}*{\%} & \multirow{1.2}*{split 1} & \multirow{1.2}*{split 2} & \multirow{1.2}*{split 3}\\
\hline
\multirow{1.2}*{Encoder} & \multirow{1.2}*{5.4} & \multirow{1.2}*{8.4} & \multirow{1.2}*{5.2}\\
\multirow{1.2}*{Residual} & \multirow{1.2}*{7.0} & \multirow{1.2}*{9.5} & \multirow{1.2}*{10.0}\\
\multirow{1.2}*{Decoder} & \multirow{1.2}*{19.8} & \multirow{1.2}*{22.3} & \multirow{1.2}*{22.53}\\
\multirow{1.2}*{Total} & \multirow{1.2}*{10.8} & \multirow{1.2}*{13.4} & \multirow{1.2}*{12.6}\\
\hline

\end{tabular}
\end{table}


\section{Conclusions}\label{sec:Conc}
By adjusting the position of the IF neuron and introducing the attention module into SNNs, we proposed a efficient yet effective multi\-step SNN model for dense stereo depth prediction, whose results outperforms the current \itshape state-of-art \upshape ANN model DDES \cite{DDES}. In addition, we proposed a new protocal, the cumulative stacking protocal, for encoding input event streams to make the MVSEC dataset suitable as input to SNNs. Experiments show that our proposed network has both high performance over ANN and very low pulse release rate, i.e., effective and efficient. The model we propose is not just usable for depth estimation, but a general model that can handle event camera regression tasks. In the next step, we hope to combine with brain-like chip hardware to truly simulate the neural system of an organism from the dataset, to the algorithm, and to the hardware level.


\bibliographystyle{IEEETran}
\bibliography{./ref}

\begin{thebibliography}{10}
\providecommand{\url}[1]{#1}
\csname url@samestyle\endcsname
\providecommand{\newblock}{\relax}
\providecommand{\bibinfo}[2]{#2}
\providecommand{\BIBentrySTDinterwordspacing}{\spaceskip=0pt\relax}
\providecommand{\BIBentryALTinterwordstretchfactor}{4}
\providecommand{\BIBentryALTinterwordspacing}{\spaceskip=\fontdimen2\font plus
\BIBentryALTinterwordstretchfactor\fontdimen3\font minus
  \fontdimen4\font\relax}
\providecommand{\BIBforeignlanguage}[2]{{%
\expandafter\ifx\csname l@#1\endcsname\relax
\typeout{** WARNING: IEEEtran.bst: No hyphenation pattern has been}%
\typeout{** loaded for the language `#1'. Using the pattern for}%
\typeout{** the default language instead.}%
\else
\language=\csname l@#1\endcsname
\fi
#2}}
\providecommand{\BIBdecl}{\relax}
\BIBdecl

\bibitem{wu2018spatio}
Y.~Wu, L.~Deng, G.~Li, J.~Zhu, and L.~Shi, ``Spatio-temporal backpropagation
  for training high-performance spiking neural networks,'' \emph{Frontiers in
  neuroscience}, vol.~12, p. 331, 2018.

\bibitem{deng2022temporal}
S.~Deng, Y.~Li, S.~Zhang, and S.~Gu, ``Temporal efficient training of spiking
  neural network via gradient re-weighting,'' \emph{arXiv preprint
  arXiv:2202.11946}, 2022.

\bibitem{wu2019direct}
Y.~Wu, L.~Deng, G.~Li, J.~Zhu, Y.~Xie, and L.~Shi, ``Direct training for
  spiking neural networks: Faster, larger, better,'' in \emph{Proceedings of
  the AAAI Conference on Artificial Intelligence}, vol.~33, no.~01, 2019, pp.
  1311--1318.

\bibitem{guiji}
G.~Debat, T.~Chauhan, B.~R. Cottereau, T.~Masquelier, M.~Paindavoine, and
  R.~Baures, ``Event-based trajectory prediction using spiking neural
  networks,'' \emph{Frontiers in computational neuroscience}, p.~47, 2021.

\bibitem{guangliu}
C.~Lee, A.~K. Kosta, A.~Z. Zhu, K.~Chaney, K.~Daniilidis, and K.~Roy,
  ``Spike-flownet: event-based optical flow estimation with energy-efficient
  hybrid neural networks,'' in \emph{European Conference on Computer
  Vision}.\hskip 1em plus 0.5em minus 0.4em\relax Springer, 2020, pp. 366--382.

\bibitem{jiaosudu}
M.~Gehrig, S.~B. Shrestha, D.~Mouritzen, and D.~Scaramuzza, ``Event-based
  angular velocity regression with spiking networks,'' in \emph{2020 IEEE
  International Conference on Robotics and Automation (ICRA)}.\hskip 1em plus
  0.5em minus 0.4em\relax IEEE, 2020, pp. 4195--4202.

\bibitem{EventSurvey}
G.~Gallego, T.~Delbr{\"u}ck, G.~Orchard, C.~Bartolozzi, B.~Taba, A.~Censi,
  S.~Leutenegger, A.~J. Davison, J.~Conradt, K.~Daniilidis \emph{et~al.},
  ``Event-based vision: A survey,'' \emph{IEEE transactions on pattern analysis
  and machine intelligence}, vol.~44, no.~1, pp. 154--180, 2020.

\bibitem{GarrickOrchard2015HFirstAt}
G.~Orchard, C.~Meyer, R.~Etienne-Cummings, C.~Posch, N.~Thakor, and
  R.~Benosman, ``Hfirst: A temporal approach to object recognition,''
  \emph{IEEE transactions on pattern analysis and machine intelligence},
  vol.~37, no.~10, pp. 2028--2040, 2015.

\bibitem{Connor2013}
P.~O'Connor, D.~Neil, S.-C. Liu, T.~Delbruck, and M.~Pfeiffer, ``Real-time
  classification and sensor fusion with a spiking deep belief network,''
  \emph{Frontiers in neuroscience}, vol.~7, p. 178, 2013.

\bibitem{SpikeFlowNet}
C.~Lee, A.~K. Kosta, A.~Z. Zhu, K.~Chaney, K.~Daniilidis, and K.~Roy,
  ``Spike-flownet: Event-based optical flow estimation with energy-efficient
  hybrid neural networks,'' in \emph{Computer Vision--ECCV 2020: 16th European
  Conference, Glasgow, UK, August 23--28, 2020, Proceedings, Part XXIX}, 2020,
  pp. 366--382.

\bibitem{StereoSpike}
U.~Ran{\c{c}}on, J.~Cuadrado-Anibarro, B.~R. Cottereau, and T.~Masquelier,
  ``Stereospike: Depth learning with a spiking neural network,'' \emph{arXiv
  preprint arXiv:2109.13751}, 2021.

\bibitem{MS-ResNet}
Y.~Hu, Y.~Wu, L.~Deng, and G.~Li, ``Advancing residual learning towards
  powerful deep spiking neural networks,'' \emph{arXiv preprint
  arXiv:2112.08954}, 2021.

\bibitem{ManYao2022AttentionSN}
M.~Yao, G.~Zhao, H.~Zhang, Y.~Hu, L.~Deng, Y.~Tian, B.~Xu, and G.~Li,
  ``Attention spiking neural networks,'' \emph{arXiv preprint
  arXiv:2209.13929}, 2022.

\bibitem{AlexZihaoZhu2018TheMV}
A.~Z. Zhu, D.~Thakur, T.~{\"O}zaslan, B.~Pfrommer, V.~Kumar, and K.~Daniilidis,
  ``The multivehicle stereo event camera dataset: An event camera dataset for
  3d perception,'' \emph{IEEE Robotics and Automation Letters}, vol.~3, no.~3,
  pp. 2032--2039, 2018.

\bibitem{LinWang2018EventbasedHD}
L.~Wang, Y.-S. Ho, K.-J. Yoon \emph{et~al.}, ``Event-based high dynamic range
  image and very high frame rate video generation using conditional generative
  adversarial networks,'' in \emph{Proceedings of the IEEE/CVF Conference on
  Computer Vision and Pattern Recognition}, 2019, pp. 10\,081--10\,090.

\bibitem{OlafRonneberger2015UNetCN}
O.~Ronneberger, P.~Fischer, and T.~Brox, ``U-net: Convolutional networks for
  biomedical image segmentation,'' in \emph{International Conference on Medical
  image computing and computer-assisted intervention}.\hskip 1em plus 0.5em
  minus 0.4em\relax Springer, 2015, pp. 234--241.

\bibitem{SanghyunWoo2018CBAMCB}
S.~Woo, J.~Park, J.-Y. Lee, and I.~S. Kweon, ``Cbam: Convolutional block
  attention module,'' in \emph{Proceedings of the European conference on
  computer vision (ECCV)}, 2018, pp. 3--19.

\bibitem{ReneRanftl2020TowardsRM}
R.~Ranftl, K.~Lasinger, D.~Hafner, K.~Schindler, and V.~Koltun, ``Towards
  robust monocular depth estimation: Mixing datasets for zero-shot
  cross-dataset transfer,'' \emph{IEEE transactions on pattern analysis and
  machine intelligence}, 2020.

\bibitem{SpikingJelly}
W.~Fang, Y.~Chen, J.~Ding, D.~Chen, Z.~Yu, H.~Zhou, Y.~Tian, and other
  contributors, ``Spikingjelly,''
  \url{https://github.com/fangwei123456/spikingjelly}, 2020, accessed:
  2022-11-18.

\bibitem{tulyakov2019learning}
S.~Tulyakov, F.~Fleuret, M.~Kiefel, P.~Gehler, and M.~Hirsch, ``Learning an
  event sequence embedding for dense event-based deep stereo,'' in
  \emph{Proceedings of the IEEE/CVF International Conference on Computer
  Vision}, 2019, pp. 1527--1537.

\bibitem{DDES}
A.~Z. Zhu, L.~Yuan, K.~Chaney, and K.~Daniilidis, ``Unsupervised event-based
  learning of optical flow, depth, and egomotion,'' in \emph{2019 IEEE/CVF
  Conference on Computer Vision and Pattern Recognition (CVPR)}.\hskip 1em plus
  0.5em minus 0.4em\relax IEEE Computer Society, 2019, pp. 989--997.

\bibitem{TESE}
A.~Z. Zhu, Y.~Chen, and K.~Daniilidis, ``Realtime time synchronized event-based
  stereo,'' in \emph{Proceedings of the European Conference on Computer Vision
  (ECCV)}, 2018, pp. 433--447.

\bibitem{CopNet}
E.~Piatkowska, J.~Kogler, N.~Belbachir, and M.~Gelautz, ``Improved cooperative
  stereo matching for dynamic vision sensors with ground truth evaluation,'' in
  \emph{2017 IEEE Conference on Computer Vision and Pattern Recognition
  Workshops (CVPRW)}.\hskip 1em plus 0.5em minus 0.4em\relax IEEE, 2017, pp.
  370--377.

\end{thebibliography}

\clearpage

\end{document}